\documentclass[runningheads]{llncs}

\usepackage{eccv}
\usepackage{eccvabbrv}

\usepackage{graphicx}
\usepackage{booktabs}

\usepackage[accsupp]{axessibility}  

\usepackage{hyperref}

\usepackage{orcidlink}

\definecolor{highlight}{rgb}{1, 0.94, 0.87}

\newcommand{\tf}[1]{\mathbf{#1}}

\definecolor{Note_color}{rgb}{0.0, 0.0, 1.0}

\usepackage{pifont}
\newcommand\mynuma[1]{\ifcase#1 \or \ding{172}\or \ding{173}\or
  \ding{174}\or \ding{175}\or \ding{176}\or \ding{177}%
  \or \ding{178}\or \ding{179}\or \ding{180}\or \ding{181}\else *\fi\relax}

\newcommand\mynumb[1]{\ifcase#1 \or \ding{182}\or \ding{183}\or
  \ding{184}\or \ding{185}\or \ding{186}\or \ding{187}%
  \or \ding{188}\or \ding{189}\or \ding{190}\or \ding{191}\else *\fi\relax}

\usepackage[flushleft]{threeparttable} %
\usepackage{multirow}
\usepackage{bbding}
\usepackage[textsize=tiny]{todonotes}
\usepackage{soul}
\usepackage{xcolor}

\definecolor{myBlue}{HTML}{6C8EBF}
\usepackage{tikz}
\usepackage{pifont}
\usepackage{fontawesome}
\usepackage{bm}

\usepackage{wrapfig}

\usepackage{color, colortbl}
\usepackage{skak}

\definecolor{gray}{gray}{0.9}
\definecolor{highlight}{rgb}{1, 0.94, 0.87}

\newcommand{\METHOD}{Omni-Recon}

\begin{document}

\title{\METHOD{}: Harnessing Image-based Rendering for General-Purpose Neural Radiance Fields}

\titlerunning{\METHOD{}}

\author{Yonggan Fu \and
Huaizhi Qu \and
Zhifan Ye \and 
Chaojian Li \and \\
Kevin Zhao \and
Yingyan (Celine) Lin}

\institute{Georgia Institute of Technology \\
\email{\{yfu314,zye327,cli851,kzhao14,celine.lin\}@gatech.edu}
}

\authorrunning{Y. Fu et al.}

\maketitle

\begin{abstract}

Recent breakthroughs in Neural Radiance Fields (NeRFs) have sparked significant demand for their integration into real-world 3D applications. However, the varied functionalities required by different 3D applications often necessitate diverse NeRF models with various pipelines, leading to tedious NeRF training for each target task and cumbersome trial-and-error experiments. Drawing inspiration from the generalization capability and adaptability of emerging foundation models, our work aims to develop one general-purpose NeRF for handling diverse 3D tasks. We achieve this by proposing a framework called \METHOD{}, which is capable of (1) generalizable 3D reconstruction and zero-shot multitask scene understanding, and (2) adaptability to diverse downstream 3D applications such as real-time rendering and scene editing. 
Our key insight is that an image-based rendering pipeline, with accurate geometry and appearance estimation, can lift 2D image features into their 3D counterparts, thus extending widely explored 2D tasks to the 3D world in a generalizable manner. Specifically, our \METHOD{} features a general-purpose NeRF model using image-based rendering with two decoupled branches: one complex transformer-based branch that progressively fuses geometry and appearance features for accurate geometry estimation, and one lightweight branch for predicting blending weights of source views. This design achieves state-of-the-art (SOTA) generalizable 3D surface reconstruction quality with blending weights reusable across diverse tasks for zero-shot multitask scene understanding. In addition, it can enable real-time rendering after baking the complex geometry branch into meshes, swift adaptation to achieve SOTA generalizable 3D understanding performance, and seamless integration with 2D diffusion models for text-guided 3D editing.
Our code is available at: \url{https://github.com/GATECH-EIC/Omni-Recon}.


\end{abstract}

\section{Introduction}
\label{sec:intro}

Neural Radiance Fields (NeRFs)~\cite{mildenhall2020nerf} have garnered significant attention and hold promise to become crucial enablers for emerging 3D applications. However, deploying NeRFs in real-world scenarios still requires substantial effort, as different 3D applications often demand distinct NeRF models with varied pipelines, leading to tedious NeRF training for each target task and cumbersome trial-and-error experiments. For instance, while both cross-scene generalization and real-time rendering are desirable in the real world, generalizable NeRFs for instant reconstruction~\cite{wang2021ibrnet, chen2021mvsnerf, liu2022neural} and real-time NeRFs using mesh-based rasterization~\cite{chen2022mobilenerf, yariv2023bakedsdf, tang2023delicate} typically feature diverse pipelines, making it hard to simultaneously satisfy the two requirements. Additionally, understanding new scene properties~\cite{zhi2021place, liu2023semantic, zhang2023beyond} necessitates training new NeRF models, which is not scalable given the increasing number of scene properties of interest.

In parallel, the emergence of foundation models has driven transformative progress in real-world artificial intelligence applications. Inspired by this exciting trend, there has been a growing interest in integrating NeRFs into this realm by developing general-purpose NeRFs for handling various 3D tasks. Specifically, similar to the generalization capability and adaptability of foundation models, we expect general-purpose NeRFs to possess three essential capabilities, including \underline{(1)} generalizable 3D reconstruction for instant surface reconstruction and novel view synthesis, \underline{(2)} zero-shot multitask scene understanding of various scene properties, e.g., semantics and edges, and \underline{(3)} adaptability to diverse downstream tasks such as real-time rendering and scene editing. Achieving these capabilities is promising to enable users to effortlessly start from these general-purpose NeRFs when facing a new 3D application, significantly reducing the need for training and experimenting with various NeRF pipelines.

Despite the promise, existing NeRF pipelines have not yet realized the mentioned capabilities. Firstly, per-scene optimized NeRFs~\cite{mildenhall2021nerf,barron2021mip,barron2022mip,yariv2023bakedsdf,hedman2021baking,hu2022efficientnerf,garbin2021fastnerf} rely on costly per-scene training, thus precluding generalization capabilities. Secondly, although generalizable NeRF designs~\cite{yu2021pixelnerf,wang2021ibrnet,trevithick2021grf,reizenstein2021common,wang2022attention,chen2021mvsnerf,xu2022point,liu2022neural} can achieve cross-scene generalization, their huge computational complexity limits their suitability for application scenarios where real-time rendering is crucial. Some generalizable surface reconstruction methods~\cite{ren2023volrecon,liang2024retr} can support mesh extraction, but an expensive volumetric rendering process, independent of the extracted mesh, is still necessary for rendering novel views, leaving the demand for real-time rendering unmet. Furthermore, their potential for zero-shot scene understanding and scene editing remains unexplored.

Given the significant potential and associated challenges, our work aims to advance the integration of NeRFs into the realm of foundation models by developing general-purpose NeRFs. To achieve this goal, we propose a framework called \METHOD{}, which is an image-based rendering pipeline that realizes all the mentioned capabilities. The key insight leveraged by our \METHOD{} is that an image-based rendering pipeline, when equipped with accurate geometry and appearance estimation, can lift 2D image features into their 3D counterparts, thus extending widely explored 2D tasks to the 3D world in a generalizable manner. Specifically, our contributions are summarized as follows:

\begin{itemize}
    \item We introduce \METHOD{}, a general-purpose NeRF pipeline, which possesses both decent generalization capabilities, including generalizable 3D reconstruction and zero-shot multitask scene understanding, and adaptability to diverse downstream 3D applications.

    \item Our \METHOD{} features a general-purpose NeRF model backbone using image-based rendering with two decoupled branches: one complex transformer-based geometry branch that progressively fuses geometry and appearance features to accurately predict a signed distance function (SDF), and one lightweight appearance branch for predicting blending weights of source views from the target scene. The advantages of this design include: \underline{(1)} ensuring accurate geometry estimation for high-quality 3D reconstruction, and \underline{(2)} enabling real-time rendering after baking the complex geometry branch into meshes and using the lightweight appearance branch as a shader.

    \item We intriguingly find that a properly trained image-based rendering pipeline can directly lift 2D monocular priors to 3D novel views by reusing the blending weights predicted by the appearance branch and leverage this to enable generalizable, zero-shot, and multitask scene understanding.

    \item We demonstrate the adaptability of \METHOD{} by extending it to diverse downstream tasks, including: \underline{(1)} a Nvdiffrast-based~\cite{laine2020modular} real-time rendering pipeline using the extracted mesh,  \underline{(2)} parameter-efficient tuning (PET) of \METHOD{} on downstream generalizable scene understanding tasks, and  \underline{(3)} a new text-guided scene editing scheme by iteratively editing and reconstructing the source images from a 3D scene using 2D diffusion models.

    \item Extensive experiments demonstrate that our \METHOD{} can achieve state-of-the-art (SOTA) generalizable 3D surface reconstruction and scene understanding performance. Additionally, it can support real-time rendering of a new scene after rapid adaptation, along with easy-to-use 3D scene editing.
    
\end{itemize}

Our \METHOD{} underscores the potentially wide usage of image-based rendering pipelines in diverse real-world 3D applications, which may have been overlooked by previous works. We believe that the insights we provide could ignite future innovations in more advanced generalizable rendering pipelines.

\vspace{-0.5em}
\section{Related Work}
\vspace{-0.5em}
\label{sec:related-work}

\noindent \textbf{Neural Radiance Fields.}
NeRFs~\cite{mildenhall2021nerf}, which represent the target scene as a continuous volume with density and view-dependent color, have garnered increasing attention thanks to their SOTA rendering quality when compared to prior approaches using explicit neural representations~\cite{hedman2018,thies2019neural,lombardi19, seitz99}. Subsequent works further improve NeRFs from various angles, such as enhancing their rendering quality~\cite{barron2021mip, barron2022mip, verbin2022ref, chen2022aug}, training NeRFs with sparse views~\cite{xu2022sinnerf,niemeyer2022regnerf,deng2022depth,jain2021putting,xu2022neurallift, truong2022sparf}, extending NeRFs to large-scale scenes~\cite{tancik2022block,xiangli2022bungeenerf,turki2022mega,xiangli2021citynerf,zhang2022nerfusion}, or applying NeRFs to other tasks, such as generative modeling~\cite{chan2020pi, schwarz2020graf, wang2022clip, niemeyer2021giraffe}, dynamic scenes~\cite{li2021, ost2020neural, pumarola2021d,cao2023hexplane}, or scene understanding~\cite{kundu2022panoptic,zhi2021place,vora2021nesf,fu2022panoptic}.

\noindent \textbf{Generalizable NeRFs.} Vanilla NeRFs rely on per-scene optimization, which limits their ability to perform instant reconstruction. To address this limitation, generalizable NeRF designs~\cite{yu2021pixelnerf, wang2021ibrnet, trevithick2021grf, reizenstein2021common, wang2022attention, chen2021mvsnerf, xu2022point, liu2022neural, johari2022geonerf, chibane2021stereo} achieve cross-scene generalization by conditioning vanilla NeRFs on the source views from the target new scene. Specifically,~\cite{yu2021pixelnerf, wang2021ibrnet, trevithick2021grf, wang2022attention} take the extracted scene features from the source views as inputs to reconstruct the radiance field via a one-shot forward pass.
Subsequent works further improve cross-scene generalization capabilities by enhancing scene geometry estimation~\cite{chen2021mvsnerf, johari2022geonerf,liu2022neural} and by introducing attention mechanisms and mixture-of-experts into NeRF backbone design~\cite{wang2022attention,cong2023enhancing,liu2023semantic}.
Despite the promise of generalizable NeRFs, they still fall short in terms of rendering efficiency due to their huge computational complexity, which entails costly inference for each sampled point along the ray.
While some generalizable surface reconstruction methods~\cite{ren2023volrecon,liang2024retr} can support mesh extraction, a costly volumetric rendering process, independent of the extracted mesh, is still necessary for rendering novel views. This poses a hindrance to their deployment in 3D applications that require real-time rendering.
Furthermore, their potential for zero-shot scene understanding and scene editing remains unexplored.

\noindent \textbf{Efficient scene representations.}
Various scene representations have been proposed to accelerate NeRFs' training and inference process. To speed up NeRF training,~\cite{sun2022direct,fridovich2022plenoxels,chen2023factor,muller2022instant,chen2022tensorf} integrate explicit neural representations with volumetric rendering. In parallel, to speed up NeRF inference/rendering, ~\cite{lindell2021autoint,rebain21,reiser2021kilonerf} attempt to reduce the complexity of the multi-layer perceptron (MLP) model in NeRF.
In addition,
\cite{garbin2021fastnerf,yu2021plenoctrees,hedman2021baking,hu2022efficientnerf,liu2020neural,schwarz2022voxgraf,reiser2021kilonerf} explore the free space in a target 3D scene to improve the sampling efficiency and thus reduce unnecessary MLP inference.
Motivated by the insight of deferred rendering~\cite{thies2019deferred},~\cite{hedman2021baking, chen2022mobilenerf,yariv2023bakedsdf,tang2023delicate} pre-compute scene features offline and thus require less computation at run-time. In particular,
\cite{chen2022mobilenerf, yariv2023bakedsdf,tang2023delicate} represents NeRF as a set of textured polygons, where the textures can be pre-computed and thus the NeRF pipeline can be efficiently executed in modern graphics hardware.
However, these efficient scene representations still necessitate costly per-scene optimization, lacking generalization capabilities.

\vspace{-0.7em}
\section{\METHOD{}: The General-Purpose NeRF Backbone}
\vspace{-0.3em}
\label{sec:backbone}

To develop general-purpose NeRFs for diverse 3D tasks, in this section, we design a general-purpose NeRF backbone and demonstrate its SOTA generalizable 3D reconstruction capability. We will leverage this backbone to enable zero-shot multitask scene understanding in Sec.~\ref{sec:scene_understanding} and adapt it to diverse downstream 3D tasks, including real-time rendering, PET for 3D scene understanding, and text-guided scene editing in Sec.~\ref{sec:downstream}.

\vspace{-0.7em}
\subsection{Backbone Design Overview}
\vspace{-0.3em}
\label{sec:backbone_overview}

\textbf{Design principles.}  
Our design is driven by the following principles: \underline{(1)} Similar to foundation models, the ability to instantly reconstruct a new scene with generalizable capabilities is essential. Therefore, we leverage image-based rendering~\cite{wang2021ibrnet, chen2021mvsnerf, liu2022neural}, which lifts 2D image features into their 3D counterparts, as our backbone design and uncover their decent capabilities in diverse 3D applications, which may have been overlooked by previous works.
\underline{(2)} Considering a wide range of 3D applications that require real-time rendering, it is highly desirable that our backbone design can be seamlessly converted into efficient 3D representations, particularly meshes, to enable real-time rendering.

\begin{figure}[t!]
\vspace{-0.5em}
\centering
\includegraphics[width=0.98\linewidth]{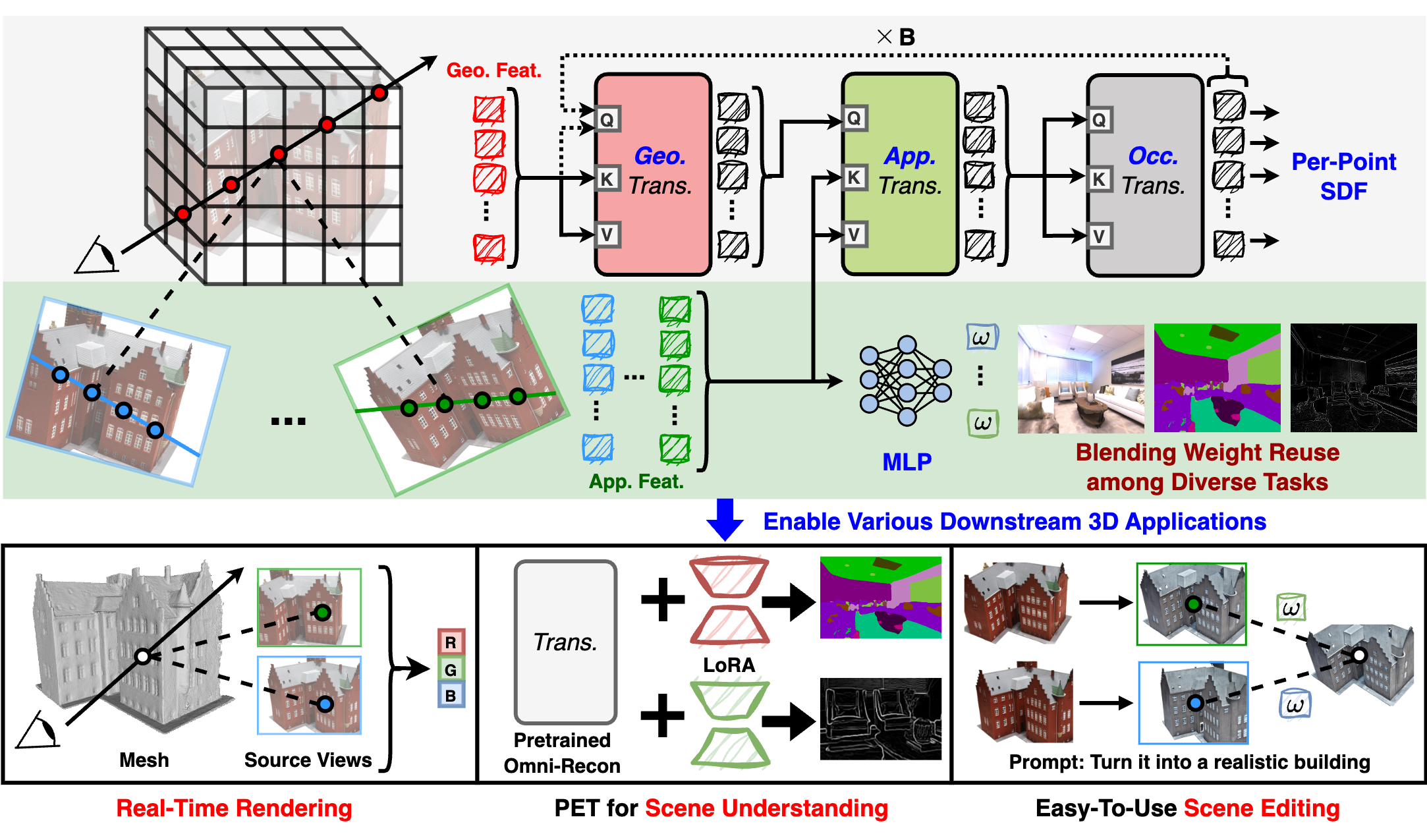}
\vspace{-1em}
\caption{An overview of our \METHOD{} framework.}
\label{fig:overview}
\vspace{-2em}
\end{figure}

\noindent \textbf{Our backbone design.} As shown in Fig.~\ref{fig:overview}, the backbone model in \METHOD{} leverages image-based rendering with two decoupled branches: one complex transformer-based geometry branch that progressively fuses geometry and appearance features to accurately predict an SDF, and one lightweight appearance branch for estimating blending weights of source views. Specifically, the progressive feature fusion in the geometry branch empowers accurate geometry estimation for high-quality 3D reconstruction, and the decoupling between geometry and appearance enables real-time rendering after baking the complex geometry branch into meshes, with the lightweight appearance branch serving as a shader. Additionally, the nature of image-based rendering, when equipped with accurate geometry and appearance estimation, can enable a wide range of downstream 3D applications, as shown in Sec.~\ref{sec:scene_understanding} and Sec.~\ref{sec:downstream}.

\vspace{-0.7em}
\subsection{The Overall Image-Based Rendering Pipeline}
\vspace{-0.3em}
\label{sec:rendering_pipeline}
 
Following generalizable NeRFs~\cite{wang2021ibrnet, yu2021pixelnerf, chen2021mvsnerf, liu2022neural}, our NeRF backbone is conditioned on source views from the target scene. This is achieved by estimating the geometry and appearance based on the extracted features of the source views from the target scene, which is elaborated as follows.

\noindent \textbf{Acquire geometry features.} Given $N$ source views $\{I_i\}_{i=1}^N \in \mathbb{R}^{H \times W \times 3}$ from the target scene,  we utilize a CNN encoder~\cite{ren2023volrecon, liang2024retr} to extract source features $\{\tf F_i\}_{i=1}^N \in \mathbb{R}^{H \times W \times C}$ from the source views. Next, we construct a 3D feature volume $V \in \mathbb{R}^{M \times M \times M \times C}$, which serves similar roles as cost volumes for multiview stereo~\cite{yao2018mvsnet, chen2021mvsnerf}, to aggregate geometry information from different source views, following~\cite{ren2023volrecon,murez2020atlas,sun2021neuralrecon}. Specifically, we divide the bounding volume of the scene into $M^3$ voxels and project each voxel center onto $N$ source views, where the mean and variance of $N$ source features are computed and concatenated as the corresponding voxel feature. We then leverage a 3D U-Net~\cite{cciccek20163d} to further enhance the feature volume to acquire the final geometry feature volume $V$.

\noindent \textbf{Acquire appearance features.} For each sampled 3D point $\tf p(t) = \tf o + t \tf d$ along the ray emitted from the target pixel on the novel view with camera origin $\tf o$ and direction $\tf d$, we project it on the source views to acquire the projected colors $\{\tf c_i\}_{i=1}^N$ and appearance features $\{\tf f_i\}_{i=1}^N$ using bilinear interpolation.

\noindent \textbf{Geometry and appearance estimation.} For more accurate reconstruction of the scene geometry and mitigation of the intrinsic color-density ambiguity~\cite{wang2021neus}, we estimate the per-point SDF and model the density as a function of SDF, following~\cite{wang2021neus,long2022sparseneus,ren2023volrecon}. For appearance, we estimate the per-point radiance through color blending, i.e., predicting the blending weights of the projected colors on the source views to compose the per-point color, following~\cite{wang2021ibrnet, yu2021pixelnerf, liu2022neural}. The geometry and appearance are modeled using separate branches and can be formulated as:

\vspace{-0.5em}
\begin{equation}
    s = \tf M_{sdf}(\{\tf f_i\}_{i=1}^N, V), \,\,\,\,\,\,  \{\omega_i\}_{i=1}^N  = \tf M_{color}(\{\tf f_i\}_{i=1}^N, \tf d), \label{eq:sdf}
\end{equation}

\noindent where $s$ is the SDF, $\{\omega_i\}_{i=1}^N$ are the blending weights, and $\tf M_{sdf}$ and $\tf M_{color}$ are the geometry and appearance branches, respectively. The per-point radiance can be derived as the weighted sum $\hat{\tf c} = \sum_{i=1}^N \omega_i \tf c_i $ across $N$ source views.

For SDF-based volumetric rendering, we adopt the formulation in NeuS~\cite{wang2021neus}. Specifically, the color is accumulated along the $K$ sampled points along each ray:

\vspace{-1em}
\begin{equation} 
    \hat{\tf C}=\sum_{j=1}^K T_j \alpha_j \hat{\tf c}_j, \,\,\,\,\, \alpha_j = 1- \exp (-\int_{t_j}^{t_{j+1}} \rho(t) dt), \label{eq:volume_rendering}
\end{equation}
\vspace{-1em}

\noindent where $T_j=\prod_{n=1}^{j-1}\left(1-\alpha_n\right)$ is the accumulated transmittance and  $\rho(t)$ is the density determined by the estimated SDF, following the definition in NeuS~\cite{wang2021neus}.

\noindent \textbf{Complexities of the two branches.} When predicting the radiance of a sampled 3D point, the colors of valid projections on the source views are already close to the radiance of this 3D point. In light of this, we employ a lightweight $\tf M_{color}$, consisting of three MLP layers, for efficient blending weight prediction. In contrast, the geometry branch should be allocated with sufficient complexity for accurately estimating the occlusion effects among different sampled points along the same ray as well as that among the sampled points and the source views. We introduce the design of the geometry branch in the next subsection.

\vspace{-0.7em}
\subsection{The Proposed Transformer-based Geometry Branch}
\vspace{-0.3em}
\label{sec:geometry_branch}

\noindent \textbf{Overview.} Our geometry branch leverages transformer modules~\cite{vaswani2017attention} to progressively fuse geometry and appearance features, thus properly handling the two occlusion effects mentioned above. Specifically, for the $K$ sampled 3D points along the ray, we acquire their geometry features $\{\tf v_k\}_{k=1}^K$ from the geometry feature volume $V$, which are sequentially processed through $B$ blocks ($B$=2 in our design), each consisting of a geometry transformer, an appearance transformer, and an occlusion transformer, to produce the per-point SDF.

\noindent \textbf{The geometry transformer.} 
The geometry transformer fuses the geometry features into its inputs by using $\{\tf v_k\}_{k=1}^K$ as the key and value in a cross-attention scheme~\cite{vaswani2017attention} and performs attention across the sampled points along the same ray to model their occlusion effects. This process can be formulated as follows:

\vspace{-1em}
\begin{equation}
    \tf M_{sdf}^{geo} (\tf x, \{\tf v_k\}_{k=1}^K) = CrossAttention \, (\tf q=\tf x, \tf k=\tf v=\{\tf v_k\}_{k=1}^K),
\end{equation}
\vspace{-1em}

\noindent where $\tf x$ is the input to the geometry transformer, which is exactly the geometry features themselves for the geometry transformer in the first block, and $\tf q$, $\tf k$, and $\tf v$ are the query, key, and value of the attention scheme~\cite{vaswani2017attention}, respectively.

\noindent \textbf{The appearance transformer.} 
The appearance transformer integrates the appearance features $\{\tf f_i\}_{i=1}^N$ into the outputs of the geometry transformer, considering the potential occlusions among the sampled points and the source views that may cause invalid projections. This integration is achieved using subtraction attention~\cite{wang2022attention,zhao2021point}, which is more effective for geometric relationship reasoning~\cite{wang2022attention,zhao2021point}, with $\{\tf f_i\}_{i=1}^N$ as the key and value. Specifically, subtraction attention computes the attention scores between the input query features and each source view in the key features, which are then used to aggregate different source views in the value features into new output features with the same dimension as the input query features. This process can be formulated as follows:

\vspace{-0.5em}
\begin{equation}
    \tf M_{sdf}^{appr} (\tf x, \{\tf f_i\}_{i=1}^N) = SubAttention \, (\tf q=\tf x, \tf k=\tf v=\{\tf f_i\}_{i=1}^N),
\end{equation}
\vspace{-1em}

\noindent where $\tf x$ is the output of $\tf M_{sdf}^{geo}$. The detailed formulation of subtraction attention is provided in the supplementary materials.

\noindent \textbf{The occlusion transformer.} We further integrate an occlusion transformer that performs self-attention across sampled points along the same ray to model their occlusion effects explicitly:

\vspace{-0.5em}
\begin{equation}
    \tf M_{sdf}^{occ} (\tf x) = SelfAttention \, (\tf q=\tf k=\tf v=\tf x),
\end{equation}
\vspace{-1em}

\noindent where $\tf x$ is the output of $\tf M_{sdf}^{appr}$. We apply another MLP on top of the output features of the last $M_{sdf}^{occ}$ module to predict the per-point SDF $s$ in Eq.~\ref{eq:sdf}.

\vspace{-0.7em}
\subsection{Training Objectives}
\vspace{-0.3em}
\label{sec:objectives}

Our NeRF backbone is trained using two objectives: (1) a photometric loss $\mathcal{L}_{rgb}$ between the rendered images and the ground-truth images, and (2) a depth loss $\mathcal{L}_{depth}$ between the rendered depth $\hat{\tf D}=\sum_{j=1}^K T_j \alpha_j t$ and the ground-truth depth. The overall objective can be formulated as follows:

\vspace{-0.5em}
\begin{equation}
    \mathcal{L} = \mathcal{L}_{color} + \beta\mathcal{L}_{depth}, \label{eq:objective}
\end{equation}

\noindent where $\mathcal{L}_{rgb} = \frac{1}{R} \sum_{\tf r=1}^R \left\lVert
   \hat{\mathbf{C}}_{\tf r} - \mathbf{C_{r}} \right\rVert_2^2$
with $R$ denoting the number of rendered pixels, $\mathcal{L}_{depth} = \frac{1}{R} \sum_{\tf r=1}^R \left\lVert \hat{\mathbf{D}}_{\tf r} - \mathbf{D_{r}} \right\rVert_2^2$, and $\beta$ is set as 1.

\begin{table}[t]
  \centering
  \caption{Benchmark the quantitative performance in sparse view mesh reconstruction in terms of Chamfer distance (the lower, the better) on 15 testing scenes on DTU. The best scores are in \textbf{bold}, and the second-best are \underline{underlined}.}
  \vspace{-1em}
    \resizebox{0.98\textwidth}{!}{
\begin{tabular}{ccccccccccccccccc}
\toprule
{ \textbf{Method}} & { \textbf{Mean}} & { \textbf{24}} & { \textbf{37}} & { \textbf{40}} & { \textbf{55}} & { \textbf{63}} & { \textbf{65}} & { \textbf{69}} & { \textbf{83}} & { \textbf{97}} & { \textbf{105}} & { \textbf{106}} & { \textbf{110}} & { \textbf{114}} & { \textbf{118}} & { \textbf{122}} \\ \midrule
COLMAP~\cite{schonberger2016pixelwise} & 1.52 & \textbf{0.90} & 2.89 & 1.63 & 1.08 & 2.18 & 1.94 & 1.61 & \underline{1.30} & 2.34 & 1.28 & 1.10 & 1.42 & 0.76 & 1.17 & 1.14 \\
MVSNet~\cite{yao2018mvsnet} & 1.22 & 1.05 & 2.52 & 1.71 & 1.04 & 1.45 & \textbf{1.52} & \underline{0.88} & \textbf{1.29} & 1.38 & 1.05 & \textbf{0.91} & \textbf{0.66} & 0.61 & 1.08 & 1.16 \\ \midrule
IDR~\cite{yariv2020multiview} & 3.39 & 4.01 & 6.40 & 3.52 & 1.91 & 3.96 & 2.36 & 4.85 & 1.62 & 6.37 & 5.97 & 1.23 & 4.73 & 0.91 & 1.72 & 1.26 \\
VolSDF~\cite{yariv2021volume} & 3.41 & 4.03 & 4.21 & 6.12 & 0.91 & 8.24 & 1.73 & 2.74 & 1.82 & 5.14 & 3.09 & 2.08 & 4.81 & 0.60 & 3.51 & 2.18 \\
UNISURF~\cite{oechsle2021unisurf} & 4.39 & 5.08 & 7.18 & 3.96 & 5.30 & 4.61 & 2.24 & 3.94 & 3.14 & 5.63 & 3.40 & 5.09 & 6.38 & 2.98 & 4.05 & 2.81 \\
NeuS~\cite{wang2021neus} & 4.00 & 4.57 & 4.49 & 3.97 & 4.32 & 4.63 & 1.95 & 4.68 & 3.83 & 4.15 & 2.50 & 1.52 & 6.47 & 1.26 & 5.57 & 6.11 \\ \midrule
PixelNeRF~\cite{yu2021pixelnerf} & 6.18 & 5.13 & 8.07 & 5.85 & 4.40 & 7.11 & 4.64 & 5.68 & 6.76 & 9.05 & 6.11 & 3.95 & 5.92 & 6.26 & 6.89 & 6.93 \\
IBRNet~\cite{wang2021ibrnet} & 2.32 & 2.29 & 3.70 & 2.66 & 1.83 & 3.02 & 2.83 & 1.77 & 2.28 & 2.73 & 1.96 & 1.87 & 2.13 & 1.58 & 2.05 & 2.09 \\
MVSNeRF~\cite{chen2021mvsnerf} & 2.09 & 1.96 & 3.27 & 2.54 & 1.93 & 2.57 & 2.71 & 1.82 & 1.72 & 2.29 & 1.75 & 1.72 & 1.47 & 1.29 & 2.09 & 2.26 \\ \midrule
SparseNeuS~\cite{long2022sparseneus} & 1.96 & 2.17 & 3.29 & 2.74 & 1.67 & 2.69 & 2.42 & 1.58 & 1.86 & 1.94 & 1.35 & 1.50 & 1.45 & 0.98 & 1.86 & 1.87 \\ 
VolRecon~\cite{ren2023volrecon} & 1.38 & 1.20 & 2.59 & 1.56 & 1.08 & 1.43 & 1.92 & 1.11 & 1.48 & 1.42 & 1.05 & 1.19 & 1.38 & 0.74 & 1.23 & 1.27 \\
ReTR~\cite{liang2024retr} & \underline{1.17} & 1.05 & \underline{2.31} & \textbf{1.44} & \underline{0.98} & \underline{1.18} & \textbf{1.52} & \underline{0.88} & 1.35 & \underline{1.30} & \underline{0.87} & 1.07 & \underline{0.77} & \underline{0.59} & \textbf{1.05} & \underline{1.12} \\
\textbf{\METHOD{} (Ours)} & \textbf{1.13} & \underline{0.91} & \textbf{2.13} & \underline{1.52} & \textbf{0.93} & \textbf{1.09} & \underline{1.70} & \textbf{0.84} & \textbf{1.29} & \textbf{1.20} & \textbf{0.83} & \underline{1.04} & 0.81 & \textbf{0.55} & \textbf{1.05} & \textbf{1.05} \\ \bottomrule
\end{tabular}
    }
  \vspace{-2em}
  \label{tab:sparse_recon}%
\end{table}%

\vspace{-0.7em}
\subsection{Evaluation: Generalizable 3D Reconstruction}
\vspace{-0.3em}
\label{sec:backbone_evaluation}

\noindent \textbf{Setup.} We assess the backbone of \METHOD{} by evaluating its generalizable 3D reconstruction capability. \underline{Datasets:}
Following~\cite{chen2021mvsnerf, long2022sparseneus, ren2023volrecon, liang2024retr}, we adopt the DTU dataset~\cite{aanaes2016large}, which comprises 124 different scenes and 7 distinct lighting conditions. For testing, 15 scenes are utilized, while the remaining scenes are allocated for training, following the dataset split in~\cite{long2022sparseneus, ren2023volrecon, liang2024retr}.
\underline{Training settings:}
During training, we employ $N = 4$ source views with a resolution of 640$\times$512. The ray number per batch and batch size are set to 1024 and 2, respectively, following~\cite{ren2023volrecon, liang2024retr}.
\underline{Mesh extraction:} We utilize TSDF fusion~\cite{sun2021neuralrecon, newcombe2011kinectfusion} to reconstruct scene meshes from predicted source view depths, following~\cite{ren2023volrecon, liang2024retr}.
More detailed model/training settings can be found in the supplementary materials.

\noindent \textbf{Baselines.}  We primarily benchmark against SOTA generalizable neural implicit reconstruction methods~\cite{croce2019sparse,ren2023volrecon,liang2024retr}, along with generalizable neural rendering methods~\cite{chen2021mvsnerf, yu2021pixelnerf, wang2021ibrnet}, per-scene optimization methods~\cite{yariv2020multiview, yariv2021volume, oechsle2021unisurf, wang2021neus}, and multi-view stereo methods~\cite{schonberger2016pixelwise, yao2018mvsnet}. We directly report their official results in~\cite{ren2023volrecon, liang2024retr}.

\begin{figure}[t!]
\centering
\includegraphics[width=0.98\linewidth]{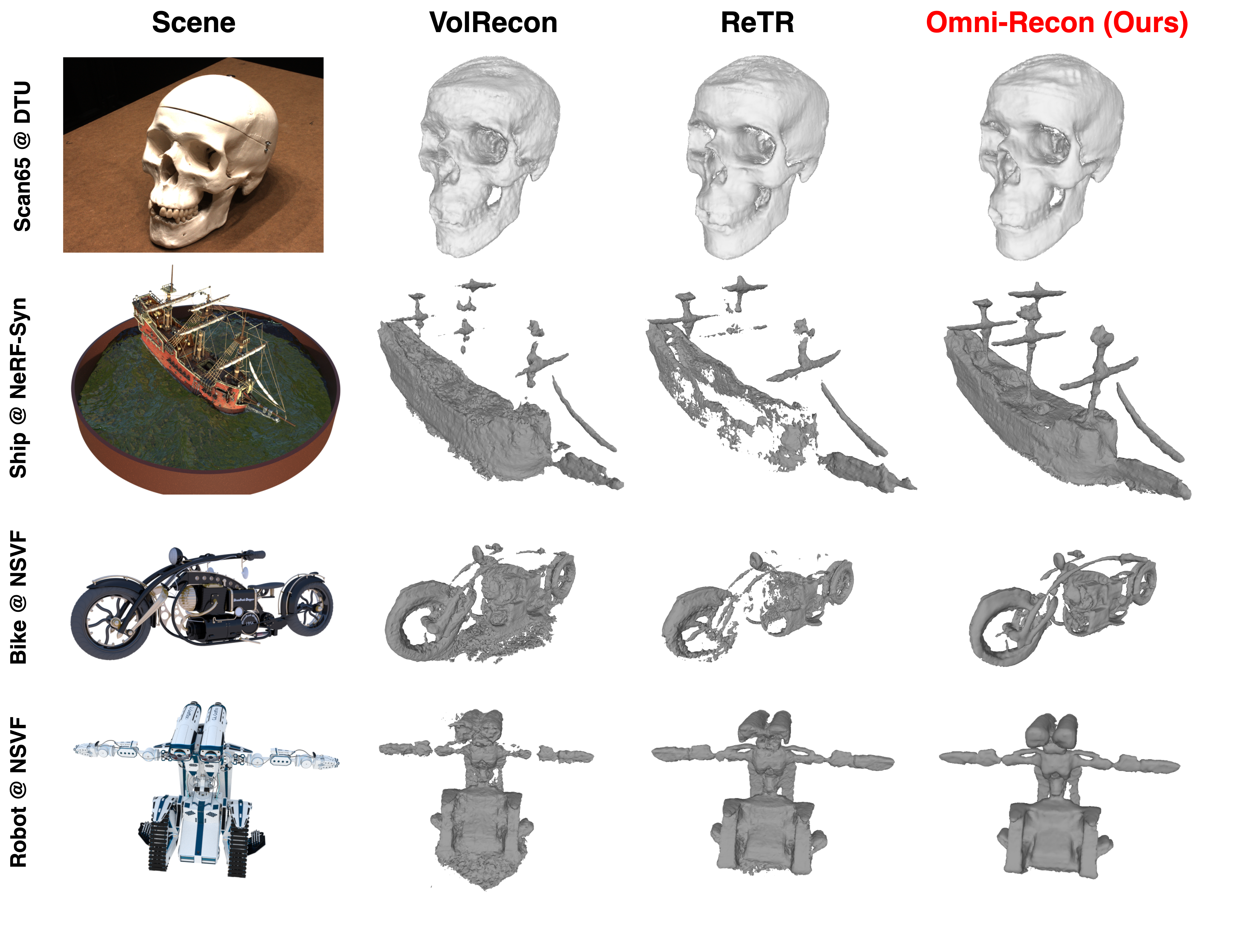}
\vspace{-2.5em}
\caption{Visualize the reconstructed mesh of our \METHOD{} and the two strongest baselines~\cite{ren2023volrecon, liang2024retr}. Row 1: Scan65 from the test scenes of DTU; Rows 2-4: Scenes from NeRF-Synthetic~\cite{mildenhall2020nerf} and NSVF-Synthetic~\cite{liu2020neural}, which present relatively challenging cases due to domain shifts w.r.t. the training scenes from DTU.}
\label{fig:recon}
\vspace{-1em}
\end{figure}

\noindent \textbf{Sparse view reconstruction.} We benchmark the performance of sparse view reconstruction using three views from each testing scene in DTU, following the setting in~\cite{croce2019sparse, ren2023volrecon, liang2024retr}. As shown in Tab.~\ref{tab:sparse_recon}, we can observe that \underline{(1)} Our \METHOD{} can achieve new SOTA reconstruction performance, with the lowest Chamfer distance averaged over all scenes. Additionally, as visualized in Fig.~\ref{fig:recon} (Row 1), our reconstructed mesh can exhibit smooth surfaces and correctly maintain the structure, e.g., the hole in the mouth. \underline{(2)} Our method achieves the best reconstruction performance in 10 out of 15 scenes, indicating its general effectiveness across diverse scenes.

\noindent \textbf{Generalization to more diverse scenes.} To demonstrate the generalization capability on more diverse scenes with larger domain shifts compared to the training scenes, we further benchmark the mesh reconstruction quality on the NeRF-Synthetic dataset~\cite{mildenhall2020nerf} and NSVF-Synthetic~\cite{liu2020neural} under a full-view reconstruction setting, i.e., the depth of each source view from each scene's training set is estimated and fused into a mesh using TSDF fusion~\cite{sun2021neuralrecon, newcombe2011kinectfusion}. As visualized in Fig.~\ref{fig:recon} (Rows 2-4), we observe that our \METHOD{} can deliver notably higher-quality meshes with better-maintained structures, fewer holes, and smoother surfaces compared to the strongest baselines ReTR~\cite{liang2024retr} and VolRecon~\cite{ren2023volrecon}. The decent generalization capability of our method is attributed to the proper decoupling and fusion of geometry and appearance features in our backbone. More visualizations and ablation studies, such as the contributions of each backbone component, are provided in the supplementary materials.

\noindent \textbf{Rendering quality.} We further benchmark the performance of generalizable novel view rendering on the test scenes from DTU. As shown in Tab.~\ref{tab:sparse_recons_psnr}, our method achieves the highest PSNR, e.g., a +0.73 and +1.74 PSNR improvement over ReTR and VolRecon, respectively. As will be demonstrated in Sec.~\ref{sec:exp_downstream}, this PSNR can be notably boosted with super-fast Nvdiffrast-based finetuning~\cite{laine2020modular}.

\begin{table}[t]
  \centering
  \caption{Benchmark the quality of novel view rendering in terms of PSNR on DTU.}
  \vspace{-1em}
    \resizebox{0.98\textwidth}{!}{
\begin{tabular}{ccccccccccccccccc}
\toprule
\textbf{Method} & \textbf{Mean.} & \textbf{24} & \textbf{37} & \textbf{40} & \textbf{55} & \textbf{63} & \textbf{65} & \textbf{69} & \textbf{83} & \textbf{97} & \textbf{105} & \textbf{106} & \textbf{110} & \textbf{114} & \textbf{118} & \textbf{122} \\ \midrule
VolRecon~\cite{ren2023volrecon} & 24.58 & 22.33 & 20.59 & 21.53 & 23.72 & 24.2 & 23.65 & 24.47 & 22.77 & 23.54 & 22.62 & 26.89 & 27.44 & 25.76 & 30.14 & 29.19 \\
ReTR~\cite{liang2024retr} & 25.59 & 24.32 & 21.84 & 23.4 & 24.56 & 26.31 & 24.5 & 24.63 & 24.3 & 24.58 & 23.85 & 27.84 & 27.97 & 26.76 & 30.03 & 28.96 \\
\textbf{\METHOD{}} & \textbf{26.32} & \textbf{24.77} & \textbf{22.33} & \textbf{23.92} & \textbf{25.56} & \textbf{26.37} & \textbf{24.75} & \textbf{25.19} & \textbf{24.94} & \textbf{24.92} & \textbf{25.06} & \textbf{28.39} & \textbf{28.63} & \textbf{28.01} & \textbf{31.49} & \textbf{30.5} \\ \bottomrule
\end{tabular}
    }
  \vspace{-1.5em}
  \label{tab:sparse_recons_psnr}%
\end{table}%
\vspace{-0.5em}
\section{\METHOD{}: Enable Zero-shot Scene Understanding}
\vspace{-0.5em}
\label{sec:scene_understanding}

To enable the understanding of various 3D scene properties, one approach is to jointly train the NeRF backbone on all target tasks. However, this approach is not scalable due to the diversity of scene properties, such as semantic segmentation, edge detection, and keypoint prediction, making it costly to cover new properties of interest. Therefore, it is highly desirable to enable zero-shot scene understanding for new tasks that were not seen during training.

\vspace{-1.3em}
\subsection{Our Strategy: The Predict-then-Blend Strategy}
\vspace{-0.3em}
\label{sec:blending_weight_reuse}

\noindent \textbf{Our hypothesis.}
We hypothesize that with accurate geometry and appearance estimation, the blending weights initially learned for radiance can be repurposed for other scene understanding tasks. In essence, just as blending source colors yields radiance for each sampled 3D point, blending the properties of source views using the same weights can yield other properties.
This hypothesis stems from the observation that scene appearance is closely related to other scene properties and thus regions with similar appearance are likely to share similar scene properties, such as semantics and edges. As such, the blending weights learned for appearance can also indicate the weighting for other scene properties.

\noindent \textbf{Our method.} Motivated by this hypothesis, our \METHOD{} employs a predict-then-blend strategy to enable zero-shot multitask scene understanding by lifting 2D monocular priors to 3D novel views. Specifically, considering the prevalence of 2D monocular vision models, our method first utilizes pretrained 2D models to generate predictions (e.g., logits in semantic segmentation) on each source view, denoted as $\{\tf P_i\}_{i=1}^N$. Then, the property of each sampled point can be obtained by blending their projected predictions on source views $\{\tf p_i\}_{i=1}^N$, reusing the RGB blending weights, i.e., $\hat{\tf p} = \sum_{i=1}^N \omega_i \tf p_i $. The final pixel-wise prediction can be derived using volumetric rendering similar to RGB reconstruction in Eq.~\ref{eq:volume_rendering}.

\begin{figure}[t!]
\centering
\includegraphics[width=0.98\linewidth]{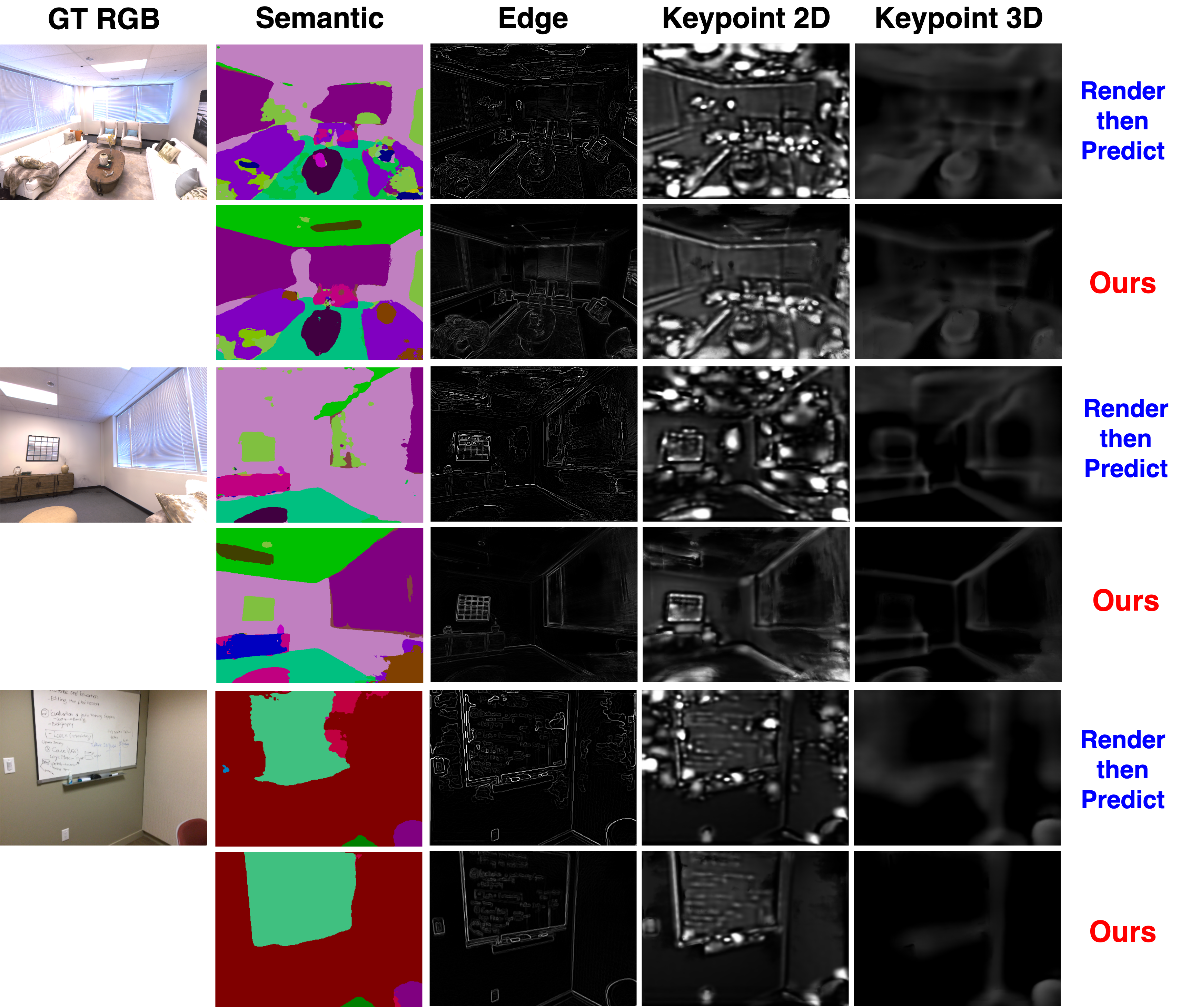}
\vspace{-0.7em}
\caption{Benchmark our predict-then-blend strategy with the render-then-predict baseline across different scene understanding tasks on Replica~\cite{straub2019replica} and ScanNet~\cite{dai2017scannet}.}
\label{fig:zero_shot_all_task}
\vspace{-2em}
\end{figure}

\vspace{-0.5em}
\subsection{Evaluation: Zero-shot Multitask Scene Understanding}
\label{sec:zero_shot_exp}

\noindent \textbf{Setup.} To evaluate the zero-shot scene understanding capabilities of our model, we adopt two indoor 3D scene datasets: Replica~\cite{straub2019replica} and ScanNet~\cite{dai2017scannet}, following the dataset splits in~\cite{zhi2021place} and~\cite{liu2023semantic}, respectively. Specifically, we focus on four scene understanding tasks: semantic segmentation, edge detection, and two keypoint detection tasks defined in~\cite{zamir2018taskonomy}. For these tasks, the 2D monocular priors are provided by~\cite{Ranftl2021},~\cite{canny1986computational}, and~\cite{midLevelReps2018}, respectively. We employ mIoU for the semantic segmentation task and $\ell_1$ error for the other three tasks as default metrics.

\noindent \textbf{Baseline.} An intuitive way to achieve zero-shot 3D scene understanding is to leverage existing 2D monocular models to predict the scene properties on top of the rendered images, a process referred to as a render-then-predict pipeline. We benchmark this strategy against our proposed predict-then-blend strategy.

\begin{wraptable}{r}{0.5\textwidth}
  \vspace{-3.5em}
  \centering
  \caption{Benchmark the two zero-shot scene understanding strategies across tasks.}
    \resizebox{0.5\textwidth}{!}{
\begin{tabular}{cccccc}
\toprule
\textbf{Dataset} & \textbf{Method} & \textbf{$\uparrow$ Semantic} & \textbf{$\downarrow$ Edge} & \begin{tabular}[c]{@{}c@{}}\textbf{$\downarrow$ Key}\\ \textbf{Point}\end{tabular} & \begin{tabular}[c]{@{}c@{}}\textbf{$\downarrow$ Key}\\ \textbf{Point 3D}\end{tabular} \\ \midrule
\multirow{2}{*}[-7pt]{Replica~\cite{straub2019replica}} & \begin{tabular}[c]{@{}c@{}}Render-then\\ -Predict\end{tabular} & 15.64 & 0.0456 & 0.1101 & 0.0470 \\ \cmidrule{3-6}
 & \begin{tabular}[c]{@{}c@{}}\textbf{Predict-then}\\ \textbf{-Blend (Ours)}\end{tabular} & \textbf{32.11} & \textbf{0.0412} & \textbf{0.0774} & \textbf{0.0176} \\ \midrule \midrule
\multirow{2}{*}[-7pt]{ScanNet~\cite{dai2017scannet}} & \begin{tabular}[c]{@{}c@{}}Render-then\\ -Predict\end{tabular} & 41.32 & 0.0471 & 0.0568 & 0.0412 \\ \cmidrule{3-6}
 & \begin{tabular}[c]{@{}c@{}}\textbf{Predict-then}\\ \textbf{-Blend (Ours)}\end{tabular} & \textbf{61.11} & \textbf{0.0434} & \textbf{0.0424} & \textbf{0.0197}  \\ \bottomrule
\end{tabular}
    }
  \vspace{-2.5em}
  \label{tab:zero_shot}%
\end{wraptable}%

\noindent \textbf{Zero-shot scene understanding benchmark.} As shown in Tab.~\ref{tab:zero_shot} and Fig.~\ref{fig:zero_shot_all_task}, we observe that \underline{(1)} Our predict-then-blend strategy achieves decent scene understanding performance on unseen scenes and unseen tasks, which echoes our hypothesis in Sec.~\ref{sec:blending_weight_reuse} that 2D monocular priors can be elevated to 3D novel views through blending weight reuse, thanks to the high correlation between scene appearance and other properties;
\underline{(2)} Compared to the render-then-predict scheme, our method achieves notably higher quantitative accuracy, e.g., a 19.79\% higher mIoU on ScanNet, and consistently better visual effects across all tasks. We attribute this improvement to two aspects: (1) Our predict-then-blend strategy can avoid feeding noisy inputs caused by rendering errors to the 2D monocular models, thus reducing reliance on the robustness of monocular models and ensuring more accurate monocular priors; (2) The multiview information gathered by our strategy can contribute to a more accurate understanding of objects with limited observations under a monocular setting.
More experiments on other vision models like CLIP-LSeg~\cite{li2022languagedriven} are in the supplementary materials.

\vspace{-1.2em}
\section{\METHOD{}: Adaptability to Downstream 3D Tasks}
\vspace{-0.5em}
\label{sec:downstream}

In addition to the generalization capabilities, we further demonstrate the adaptability of our \METHOD{} by extending it to three different types of downstream 3D applications. These demonstrations highlight the potentially broad utility of image-based rendering pipelines, which may have been overlooked by previous works, and can be generalized to a wider range of real-world 3D applications.

\vspace{-1.2em}
\subsection{Application 1: A New Real-time Rendering Pipeline}
\vspace{-0.3em}
\label{sec:realtime_rendering}

To rapidly enable real-time rendering using \METHOD{}'s pretrained backbone, we build a real-time rendering pipeline based on Nvdiffrast~\cite{laine2020modular}. This involves baking the complex geometry branch into meshes using TSDF fusion~\cite{sun2021neuralrecon,newcombe2011kinectfusion} as discussed in Sec.~\ref{sec:backbone_evaluation}, and using the lightweight appearance branch as a shader.

\noindent \textbf{Our real-time rendering pipeline.} As illustrated in Fig.~\ref{fig:overview} (bottom left), for any novel view of a new scene, we first transform its camera pose to clip space as required by Nvdiffrast~\cite{laine2020modular}. Then, we perform a rasterization process to obtain the intersection point of each camera ray with the extracted mesh. Next, these intersection points are input into the shader to predict the color by blending their projected colors on the source views as described in Sec.~\ref{sec:rendering_pipeline}. Thanks to the blending scheme, where valid projections can provide strong appearance clues, the rendering quality is often satisfactory even without finetuning.

\noindent \textbf{Joint mesh and shader finetuning.} To further enhance the rendering quality, we can jointly finetune the mesh and the shader in a gradient-based manner using differentiable rendering enabled by Nvdiffrast~\cite{laine2020modular}. Specifically, we finetune the location of each vertex in the mesh and the parameters of the lightweight shader. Additionally, we periodically prune redundant mesh faces that cannot intersect with any camera rays in one training epoch. More details are provided in the supplementary materials.

\vspace{-1em}
\subsection{Application 2: Parameter-Efficient Tuning for 3D Understanding}
\vspace{-0.3em}
\label{sec:app_scene_understanding}

We explore the possibility of finetuning \METHOD{} for various downstream 3D tasks, utilizing a series of 3D scene understanding tasks as demonstrations. 
Specifically, we conduct PET~\cite{ding2023parameter} on top of \METHOD{}'s pretrained backbone by incorporating low-rank adapters~\cite{hu2021lora} into the linear layers within the transformer modules introduced in Sec.~\ref{sec:geometry_branch}, and only tuning these adapters while keeping the pretrained backbone frozen. It is worth noting that when evaluating the performance of the tuned model, we assess it on new scenes that have never been encountered during both training and tuning, representing a generalizable scene understanding scenario~\cite{liu2023semantic}.

\vspace{-1em}
\subsection{Application 3: Text-guided 3D Scene Editing}
\vspace{-0.3em}
\label{sec:app_scene_editing}

We target text-guided 3D scene editing, which involves editing a 3D scene based on textual instructions. To achieve this, our \METHOD{} capitalizes on the nature of image-based rendering, allowing edits made to 2D source views to be propagated to 3D novel views. Specifically, leveraging the availability of powerful 2D diffusion models, we use publicly available ones to edit the 2D source views. However, doing so naively may result in inconsistent source views, as the same textual instruction may lead to different visualizations, thus degrading reconstruction quality. Inspired by~\cite{haque2023instruct}, which iteratively edits the training images and retrains a new NeRF to achieve scene editing, we address the inconsistent editing issue by proposing an iterative editing and reconstruction pipeline.

\noindent \textbf{The iterative editing and reconstruction pipeline.} To ensure consistent 3D editing, we iterate between source view editing and reconstruction as follows: First, we edit each source view using the diffusion model InstructPix2Pix~\cite{brooks2023instructpix2pix}, which is guided by both textual instructions and the original image to maintain image fidelity. While the edited source views can faithfully follow the textual instructions, they may exhibit inconsistencies in the edited region. Next, we reconstruct each source view by conditioning on their nearby source views, as described in Sec.~\ref{sec:rendering_pipeline}. This process improves the 3D consistency among source views at the expense of reduced image fidelity and less adherence to instructions. We then iterate through the editing and reconstruction steps to progressively enhance image fidelity, 3D consistency, and adherence to instructions. The resulting source views are finally used to reconstruct the entire scene.

\begin{table}[t]
  \vspace{-0.5em}
  \centering
  \caption{Benchmark the rendering speed and rendering quality of our real-time rendering pipeline and the baselines~\cite{ren2023volrecon, liang2024retr} on DTU. The best PSNR is in \textbf{bold}, and the shortest finetuning time that our pipelines surpass the strongest baseline is \underline{underlined}.}
  \vspace{-0.5em}
    \resizebox{0.999\textwidth}{!}{
\begin{tabular}{c|c|cccccccccccccccc}
\toprule
\textbf{Method} & \textbf{FPS} & \textbf{Mean} & \textbf{24} & \textbf{37} & \textbf{40} & \textbf{55} & \textbf{63} & \textbf{65} & \textbf{69} & \textbf{83} & \textbf{97} & \textbf{105} & \textbf{106} & \textbf{110} & \textbf{114} & \textbf{118} & \textbf{122} \\ \midrule
VolRecon~\cite{ren2023volrecon} & 0.029 & 24.58 & 22.33 & 20.59 & 21.53 & 23.72 & 24.2 & 23.65 & 24.47 & 22.77 & 23.54 & 22.62 & 26.89 & 27.44 & 25.76 & 30.14 & 29.19 \\
ReTR~\cite{liang2024retr} & 0.024 & 25.59 & 24.32 & 21.84 & 23.4 & 24.56 & 26.31 & 24.5 & 24.63 & 24.3 & 24.58 & 23.85 & 27.84 & 27.97 & 26.76 & 30.03 & 28.96 \\ \midrule
Ours w/o ft. & \multirow{7}{*}{\begin{tabular}[c]{@{}c@{}}71.3 \\ (40.82) \end{tabular}} & 22.96 & 20.12 & 19.71 & 22.27 & 22.78 & 24.55 & 21.77 & 21.51 & \underline{26.72} & 22.33 & \underline{24.49} & 22.52 & 22.93 & 24.68 & 23.84 & 24.12 \\
Ours (ft. 10s) &  & \underline{25.68} & 21.42 & 21.68 & \underline{24.06} & 24.12 & \underline{28.19} & 24.10 & 23.95 & 31.65 & 24.41 & 28.15 & 25.63 & 25.85 & 26.15 & 26.82 & 26.89 \\
Ours  (ft. 20s) &  & 27.21 & 22.63 & \underline{22.92} & 25.12 & \underline{25.42} & 30.03 & \underline{25.95} & \underline{26.16} & 33.19 & \underline{26.22} & 30.36 & 27.44 & 27.04 & \underline{26.93} & 29.15 & \underline{29.65} \\
Ours  (ft. 30s) &  & 27.78 & 23.2 & 23.26 & 25.54 & 25.7 & 30.59 & 26.83 & 26.96 & 33.66 & 26.47 & 30.73 & \underline{28.14} & 27.70 & 27.1 & \underline{30.17} & 30.65 \\
Ours (ft. 1min) &  & 28.34 & \underline{24.69} & 24.11 & 25.76 & 26.05 & 30.93 & 27.66 & 27.49 & 33.68 & 27.07 & 30.97 & 28.51 & \underline{28.54} & 27.35 & 31.17 & 31.54 \\
Ours (ft. 3min) &  & 28.95 & 25.2 & 24.32 & \textbf{25.94} & 26.16 & 32.15 & 28.99 & \textbf{27.88} & 34.94 & \textbf{27.35} & 31.62 & 28.93 & 28.97 & 27.56 & 31.70 & 32.49 \\
Ours (ft. 5min) &  & \textbf{29.02} & \textbf{25.34} & \textbf{24.36} & 25.63 & \textbf{26.21} & \textbf{32.16} & \textbf{29.33} & 27.81 & \textbf{34.94} & 27.32 & \textbf{31.74} & \textbf{29.04} & \textbf{29.05} & \textbf{27.69} & \textbf{31.74} & \textbf{32.89} \\ \bottomrule
\end{tabular}
    }
  \vspace{-1.5em}
  \label{tab:realtime_rendering}%
\end{table}%

\vspace{-0.3em}
\subsection{Evaluation: Effectiveness on Different Downstream 3D Tasks}
\vspace{-0.3em}
\label{sec:exp_downstream}

\textbf{Real-time rendering.} We evaluate our real-time rendering pipeline in Sec.~\ref{sec:realtime_rendering} in terms of both rendering quality and rendering speed, measured on an NVIDIA A5000 GPU, using the DTU dataset. Specifically, we finetune the scene meshes, extracted by TSDF fusion as mentioned in Sec.~\ref{sec:backbone_evaluation}, and the shader in no more than 5 minutes and record the rendering quality at different time steps.

\noindent \textit{Observations and analysis:}
As shown in Tab.~\ref{tab:realtime_rendering}, we can observe that
\underline{(1)} Without any finetuning, the rendering pipeline already exhibits certain rendering capabilities, despite the model backbone being trained using volumetric rendering with more than one sampled point;
\underline{(2)} Our rendering pipeline can be swiftly finetuned to significantly boost the PSNR. For instance, with just a 10-second finetuning, our rendering pipeline can match the average rendering quality of the strongest baseline, ReTR~\cite{liang2024retr}; With 1-minute/2-minute finetuning, it can surpass ReTR by a +2.75 and +3.36 PSNR improvement on average, respectively;
\underline{(3)} Our rendering pipeline enables real-time rendering with 71.3 FPS, i.e., a $>$2458× speed-up over the baselines. Here, the assumption is that the feature extraction of each source view from a target scene can be done in a one-time effort and does not account for throughput. If we still consider this feature extraction latency for each rendering, the resulting 40.82 FPS still meets the real-time requirement.

These experiments indicate that \textit{given a new scene}, our \METHOD{} is promising to enable both instant and accurate reconstruction and real-time rendering.

\begin{figure}[t!]
\centering
\includegraphics[width=0.8\linewidth]{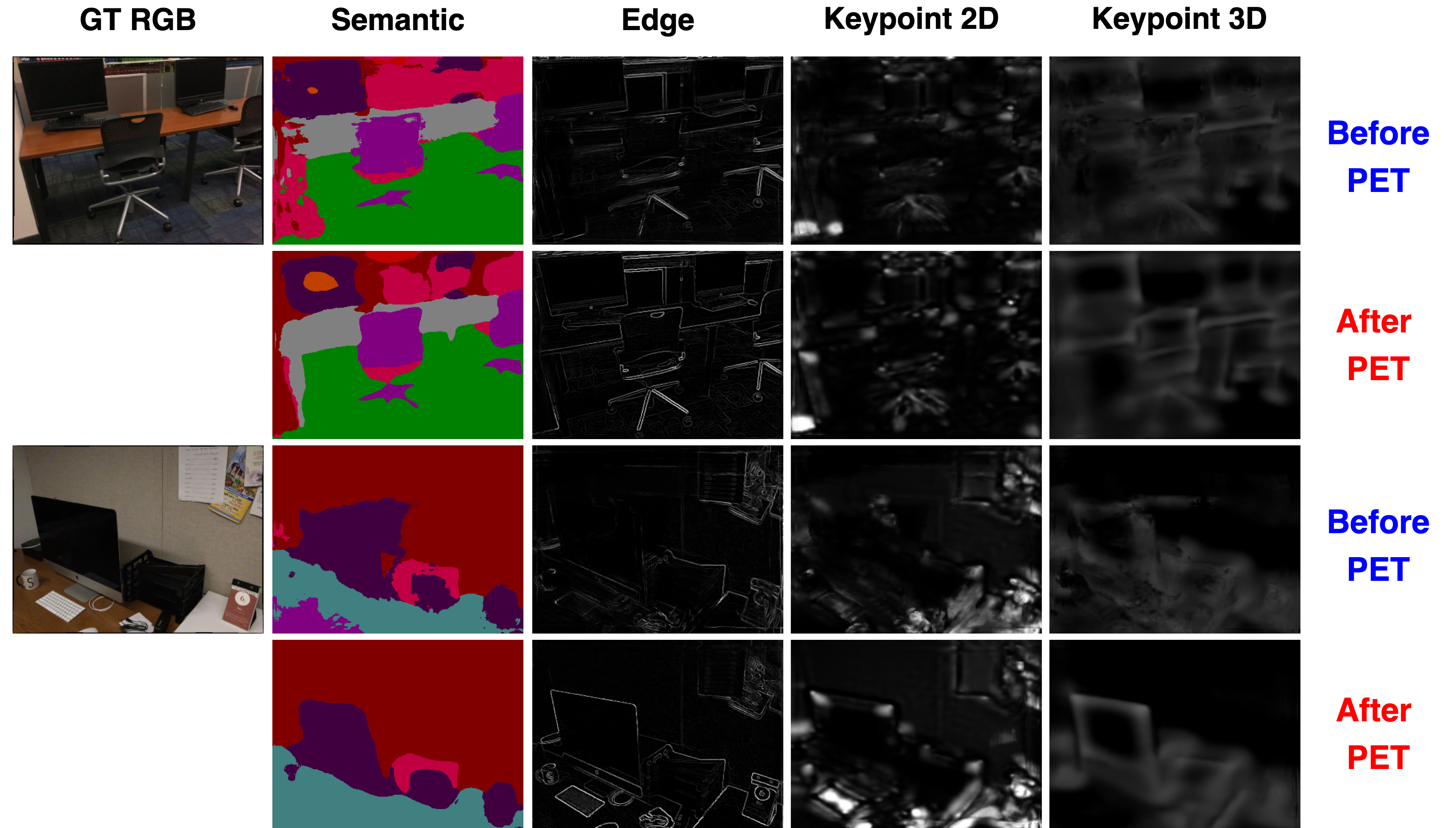}
\vspace{-0.7em}
\caption{Visualize the scene understanding performance on hard cases before/after PET.}
\label{fig:finetune}
\vspace{-1em}
\end{figure}

\begin{wraptable}{r}{0.55\textwidth}
  \vspace{-3.3em}
  \centering
  \caption{Evaluate the effectiveness of PET on \METHOD{} and benchmark with SRay~\cite{liu2023semantic}.}
    \resizebox{0.55\textwidth}{!}{
\begin{tabular}{c|cccccc}
\toprule
\textbf{Method} & \textbf{\begin{tabular}[c]{@{}c@{}}$\uparrow$ Sem. \\ mIoU\end{tabular}} & \textbf{\begin{tabular}[c]{@{}c@{}}$\uparrow$ Sem. \\ Total Acc\end{tabular}} & \textbf{\begin{tabular}[c]{@{}c@{}} $\uparrow$ Sem. \\ Avg Acc\end{tabular}} & \textbf{$\downarrow$ Edge} & \textbf{\begin{tabular}[c]{@{}c@{}}$\downarrow$ Key \\ Point\end{tabular}} & \textbf{\begin{tabular}[c]{@{}c@{}}$\downarrow$ Key \\ Point 3D\end{tabular}} \\ \midrule
SRay~\cite{liu2023semantic} & 57.15 & 78.24 & 62.55 & - & - & - \\ \midrule
\begin{tabular}[c]{@{}c@{}}Ours \\ (Zero-shot)\end{tabular} & 61.11 & 80.80 & 69.17 & 0.0434 & 0.0424 & 0.0197 \\ \midrule
\begin{tabular}[c]{@{}c@{}}Ours (PET \\ + Zero-shot)\end{tabular} & \textbf{62.35} & \textbf{81.84} & \textbf{69.89} & \textbf{0.0342} & \textbf{0.0207} & \textbf{0.0132} \\ \bottomrule
\end{tabular}
    }
  \vspace{-2.5em}
  \label{tab:pet}%
\end{wraptable}%

\noindent \textbf{PET for 3D Understanding.}  To evaluate the adaptability of our \METHOD{}, we apply PET to it for various scene understanding tasks on ScanNet~\cite{dai2017scannet}. Specifically, we finetune a LoRA adaptor~\cite{hu2021lora} for each task on a set of training scenes and evaluate on non-overlapping test scenes, representing a generalizable 3D understanding setting, following the dataset split described in~\cite{liu2023semantic}. We benchmark against the SOTA generalizable semantic segmentation work, SRay~\cite{liu2023semantic}.

\noindent \textit{Observations and analysis:} As shown in Tab.~\ref{tab:pet}, we can observe that \underline{(1)} After PET, our model outperforms SRay in terms of generalizable semantic segmentation with a +5.20\% mIoU and a +7.34\% accuracy improvement;
\underline{(2)} Despite having limited trainable parameters, PET can enhance the scene understanding performance across tasks, especially on challenging cases visualized in Fig.~\ref{fig:finetune}, indicating the adaptability of \METHOD{} to diverse downstream tasks.

\begin{figure}[t!]
\centering
\includegraphics[width=0.9\linewidth]{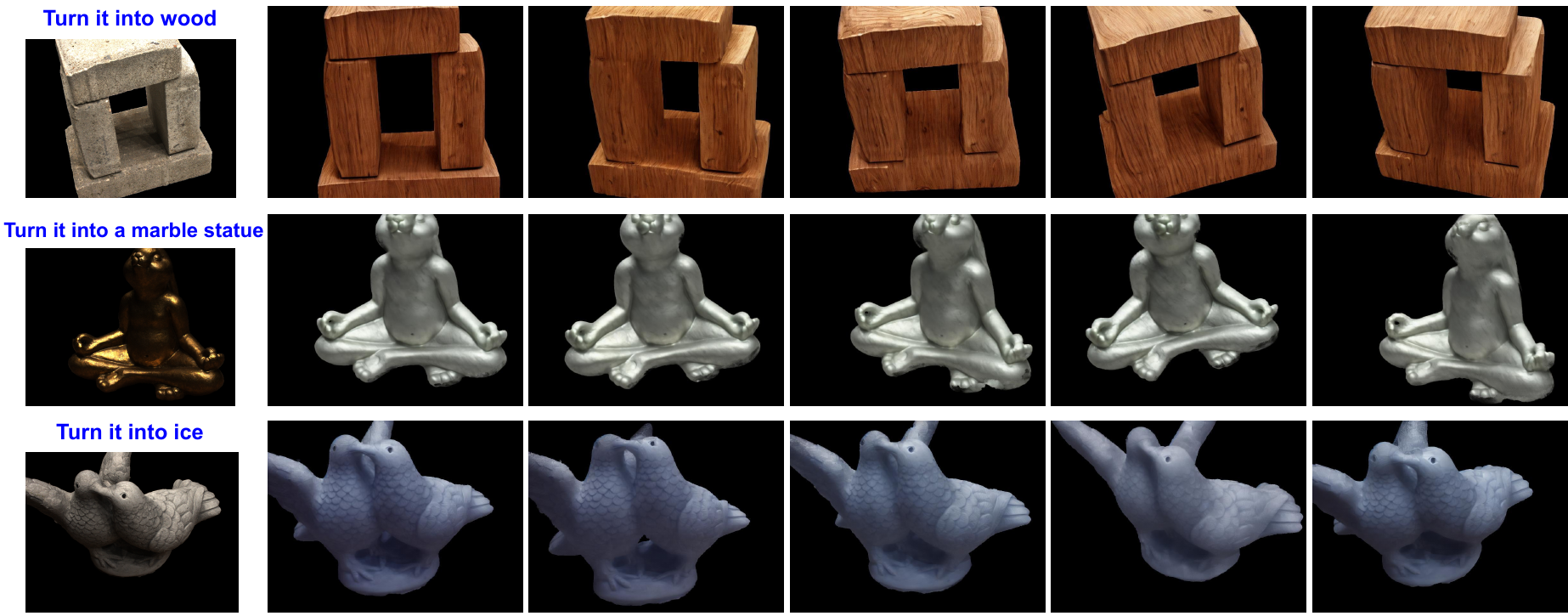}
\vspace{-0.7em}
\caption{Three text-guided scene editing examples using our pipeline in Sec.~\ref{sec:app_scene_editing}.}
\label{fig:scene_edit}
\vspace{-2em}
\end{figure}

\noindent \textbf{Text-guided scene editing.} We visualize the edited examples in Fig.~\ref{fig:scene_edit}, showcasing the ability to ensure both instruction-following and 3D consistency.
\vspace{-0.5em}
\section{Conclusion}
\vspace{-0.5em}
\label{sec:conclusion}

Motivated by the significant demand for supporting various functionalities required by different 3D applications through a unified NeRF model, our work aims to develop a general-purpose NeRF capable of handling a broad spectrum of 3D tasks. We propose a framework called \METHOD{}, which features a general-purpose NeRF model using image-based rendering with two separate branches, and demonstrate its integration with various 3D tasks. Specifically, our \METHOD{} can achieve SOTA generalizable 3D reconstruction quality, enable zero-shot multitask scene understanding, achieve SOTA scene understanding performance and real-time rendering after rapid adaptation, and support scene editing. \METHOD{} underscores the potential of image-based rendering pipelines in diverse real-world 3D applications, which may have been underestimated by prior studies, thus sparking advancements in future rendering pipelines.

\section*{Acknowledgement}
The work is supported by the National Science Foundation (NSF)
through the SCH program (Award number: 1838873) and CoCoSys, one of the seven centers in JUMP 2.0, a Semiconductor Research Corporation (SRC) program sponsored by DARPA.

\bibliographystyle{splncs04}
\bibliography{ref}
\end{document}